# AirDraw: Leveraging Smart Watch Motion Sensors for Mobile Human Computer Interactions

Danial Moazen
California State University, Northridge
Northridge, USA
gdanialq@gmail.com

Seyed A Sajjadi
California State University, Northridge
Northridge, USA
seyed.sajjadi.947@my.csun.edu

Ani Nahapetian
California State University, Northridge
Northridge, USA
ani@csun.edu

*Abstract*— **Wearable computing is one of the fastest growing technologies today. Smart watches are poised to take over at least of half the wearable devices market in the near future. Smart watch screen size, however, is a limiting factor for growth, as it restricts practical text input. On the other hand, wearable devices have some features, such as consistent user interaction and hands-free, heads-up operations, which pave the way for gesture recognition methods of text entry. This paper proposes a new text input method for smart watches, which utilizes motion sensor data and machine learning approaches to detect letters written in the air by a user. This method is less computationally intensive and less expensive when compared to computer vision approaches. It is also not affected by lighting factors, which limit computer vision solutions. The AirDraw system prototype developed to test this approach is presented. Additionally, experimental results close to 71% accuracy are presented.**

*Index Terms*— **Wearable Computing, Mobile Text Input, Smart Watches, Dynamic Time Warping (DTW).**

## I. INTRODUCTION

Wearable computing is one of the fastest growing technologies, today, with forecasts suggesting that smart watches, alone, will take over half the wearable computing market by 2018 [11]. A new approach to mobile user interface design for wearable technology is important to unlock the potential of these systems.

A limiting factor in smart watch's market growth is their small screen and interface difficulties [9]. This paper offers a new mobile text input method for smart watches, which utilizes the motion sensors on the watch as the resource for arm movement information in 3D. Using dynamic time warping (DTW) for letter classification, the system aims to extrapolate the letters the user is writing in the air. A prototype, called AirDraw, is presented and used for the system evaluation. The strong classification accuracy is presented in the experimental results.

In the area of gesture detection and recognition, most of the existing systems rely on computer vision [8]. The need for multiple cameras to make up a 3D model, the dependence on optimal lighting conditions, and computationally intensive image processing demands are some of the drawbacks of computer vision based systems for gesture recognition [8]. These shortfalls call for alternative approaches, as in our alternative leveraging motions sensors onboard wearable systems.

## II. RELATED WORK

Wearable computing is one of the successful attempts in the field of ubiquitous computing, with smart watches gaining fast popularity. Major tech companies such as Samsung, LG, Sony and recently Apple produce their own smart watches. Additionally, there are wrist based activity trackers such as those from Fitbit equipped with motion sensors able to detect, recognize, and measure several daily activities including steps and sleep patterns [12].

The user interaction difficulties, due to smart watches' small screens, have encouraged the new ideas for UI approaches. One of these approaches, proposed by Komninos and Dunlop [9], relies on taps on the touch screen as the basic text input approach. They proposed a new layout for keyboard with only 6 buttons. Some hand writing recognition approaches like Unistrokes [10] and Graffiti [6] also tried to make text input easier for users. None of these approaches are hands-free or heads-up.

Gesture recognition systems have been examined in the literature. Agrawal et al. leverage the accelerometer sensor in mobile phones to capture the information written on the air [1]. They treat each written character as a collection of strokes and classify letters according to the detected strokes. Goldberg and Richardson proposed a unistrokes approach [10], with a defined stroke for each letter. The approach requires the user to learn the shorthand for the letters to be able to with the system. Our approach uses the natural form of letter, limiting the knowledge based the user needs to interact with the system. The use of hidden Markov Model for hand gesture recognition has been considered as well [3] [4].

Abhinav Parate et al. use a wrist device to differentiate a smoking gesture from other gestures, some of which are very similar to smoking such as food intake. The project utilizes a low-power 9-axes inertial measurement unit (IMU) on a wristband. IMU provides 3D orientation of the wrist by using the data from accelerometer, gyroscope and compass [2]. Yujie Dong et al. also made a wrist based wearable device using an expensive sensor ($2,000 US) to measure the food intake [5].

## III. APPROACH

### A. AirDraw System Prototype

A prototype, called AirDraw, with hardware and software components was assembled and developed to validate the proposed approach. AirDraw consists of a smart watch and a handheld device, with communication facilitated by a Bluetooth connection between the devices. Android Wear applications were developed for the smart watch (Wear) and the handheld (Mobile). The gravity and linear acceleration of the smart watch in three axes, x, y and z, is collected and transmitted to the more resource-available handheld for processing. The hardware schema is provided in Figure III.1.

Figure III.2 illustrates the software overview of the system. The wear application housed on the smart watch filters the data to smooth the signal and to save on the data transmission cost. It also calculations the angle of the user's hand and sends that information to the handheld for manipulating the frame of reference of the x-axis and y-axis data.

The mobile application housed on the handheld processes the acceleration and angle information to obtain the rotated acceleration data that independent of the user's arm position. That data is then passed to a classifier for letter predication.

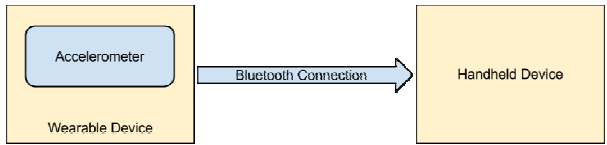

Figure III.1: Hardware schema; handheld device, wearable device and the Bluetooth connection.

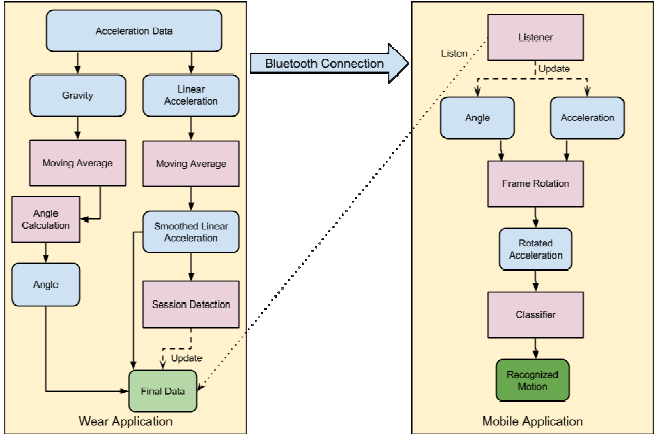

Figure III.2: Software Overview. Wear Application and Mobile Application

### B. Data Filtering

A weighted moving average filter is used to smooth the signals from sensors. Figure III.3 shows the improvement obtained from applying this smoothing algorithm. The blue line is the original acceleration signal along the z-axis obtained while drawing a circle without any smoothing filter applied. The red line shows the same signal for the very same movement with the smoothing algorithm applied.

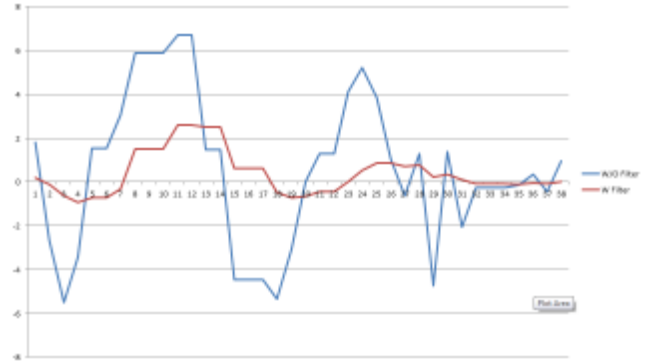

Figure III.3: Comparing the signal with and without weighted moving average filter over time. The blue line shows the acceleration signal on the z-axis while writing the letter a without any filter applied. The red line shows the same signal for the very same movement with the filter applied.

### C. Arm Angle Calculation

The angle that user's arm make with the horizon line is needed to cancel the effect of the user's arm orientation on the linear acceleration data while drawing or writing in the air. Implemented in the Wear application, the smoothed gravity signal is used to calculate the angle between that the smart watch's x-axis (or user's arm) with the horizon line. Figure III.4 shows the smart watch's fixed frame of reference and how the x-axis coincides with user's arm.

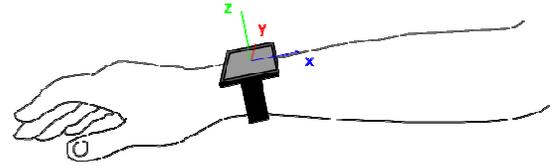

Figure III.3: Smart watch's fixed frame of reference.

The arm's orientation affects the acceleration data along the x-axis and the y-axis. To cancel its effect, the arm's angle is used to rotate the frame of reference accordingly. As a result of the rotation, up and down movements are represented in y-axis acceleration changes. Back and forth movements are represented as x-axis acceleration changes. Independent of the arm angle, left and right movements are extrapolated from the acceleration along the z-axis.

In order to calculate the angle, first we need to calculate the norm of the gravity vector. In 3D environments this is done as shown in Equation III.1.

*Equation III.1:* $\mathbf{Norm} = \sqrt{\mathbf{gx2} + \mathbf{gy2} + \mathbf{gz2}}$

With the norm, we can to normalize the gravity along each axis. Normalized gravity for each axis is calculated by dividing the gravity along each axis by the norm. The calculation for x is shown in Equation III.2.

*Equation III.2:* $gx - normal = gx / Norm$

The angle that x makes with horizon can be calculated by taking the arctangent of $g_{x\text{-normal}}$ over $g_{y\text{-normal}}$.

Figure III.4 shows the effect of the rotation on signals x and y. Both Figure III.4.a and Figure III.4.b show the acceleration data when the wearable device is moving up and down repeatedly while the device is rotating around z-axis 90 degrees. Figure III.4.a shows the data before rotation and Figure III.4.b shows the data after rotation. Note the effect of the device's orientation being canceled out along the x-axis and added to the y-axis. This addition and deletion is in the reverse direction when the device is moving back and forth.

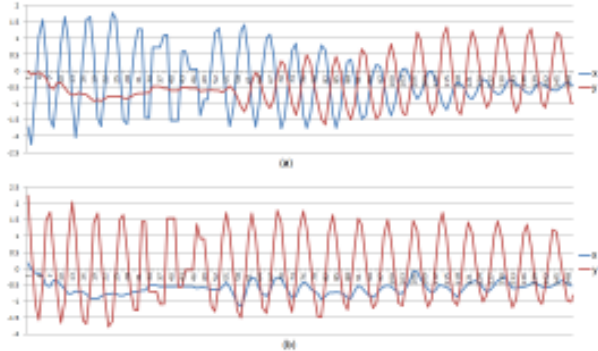

Figure III.4: Comparing the acceleration data before and after applying the rotation of the frame of reference while the device is moving up and down repeatedly and rotating around z-axis 90 degrees. 3.5.a, on the top, shows the acceleration without applying the rotation on the frame of reference and 3.5.b, on the bottom, shows the acceleration after applying the rotation on the frame of reference.

*D. Session Detection*

Communication between the devices is expensive in terms of power. To minimize that communication and the data synchronization costs, session detection is carried out by the Wear application, so that synchronization is only carried out during an active session. The time window that the user is writing a letter is referred to as session. The start of a session is signaled by the acceleration signals being greater than 1 m/s2. Once the acceleration remains less than 1 m/s2 for more than 400 milliseconds, the session will be considered complete. Table III.2 is compares the average number of data transfers between devices to write the same five words, with a continuous connection and a connection during active sessions. The data for each word is the average of five time samplings. These words are the 5 top most searched food related words in the U.S. in 2014 according to Google Trend [7].

Table III.2: Comparing the amount of data (averaged across 5 measurements) updated and transferred between devices with a continuous connection and with a connection only during active sessions. The words are the 5 top Google searched food related words in the U.S. in 2014.

| Connection Type | Air Written Words | | | | |
| --- | --- | --- | --- | --- | --- |
| | *Pizza* | *Chicken* | *Cake* | *Wine* | *Coffee* |
| Continuous | 281.4 | 305.4 | 228.8 | 201.2 | 297 |
| During Active Sessions | 186 | 198.8 | 118.2 | 118.2 | 155.8 |
| Data Transfer Savings | 33.9% | 34.9% | 48.3% | 41.5% | 47.5% |

*E. Letter Classification*

Letter classification is carried out using a supervised learning algorithm called dynamic time warping (DTW), with only one instance of the data (writing each letter once) for training. Any data that can be converted into a linear sequence can be compared with DTW, with the algorithm returning the distance between signals.

In our approach, a comparison along each axis is carried out separately. Then the total distance is calculated by adding the distances for the x-axis, y-axis and z-axis. When the user writes a letter, the acceleration data of the newly written letter is compared with the data saved in the training phase for all the letters. The letter with the minimum total distance is determined to be the intended letter.

IV. EXPERIMENTAL RESULTS

This section presents experimental results using the AirDraw implementation. All the experimentation was carried out with a single user, with the smart watch is worn on the user's dominant hand (right hand). The test for each letter was completed before starting the next letter. On average, each took 1.5 second to write.

All of the presented confusion matrices below provide the actual letter along the y-axis and the predicted letter along the x axis.

Table IV.1 illustrates the results for 5 non-similar letters (a, b, j, w and z) using the DTW algorithm. These letters were chosen as they are distinct from each other in terms of stroke shape and count. The size of the letter written on air is 12 inches by 12 inches. Each letter is written 100 times and the results are averaged.

When we decrease the range of the arm movement (size of the letter) to 6 inches by 6 inches, it is seen on table IV.2 that for some letter such as b, z, and j, the accuracy drops.

Table IV.1: Detection accuracy of DWT algorithm for 5 non-similar letters. The size of the letters is 12 inches by 12 inches. The test is done 100 times for each letter.

|  |  | Predicted | | | | |
|---|---|---|---|---|---|---|
|  |  | a | b | J | W | z |
| Actual | a | 100% | 0% | 0% | 0% | 0% |
|  | b | 5% | 95% | 0% | 0% | 0% |
|  | j | 0% | 4% | 84% | 10% | 2% |
|  | w | 0% | 10% | 16% | 74% | 0% |
|  | z | 0% | 0% | 0% | 4% | 96% |

Table IV.2: Detection accuracy of DTW algorithm for 5 non-similar letters. The size of the letters is 6 inches by 6 inches. The test is done 100 times for each letter.

|  |  | Predicted | | | | |
|---|---|---|---|---|---|---|
|  |  | a | b | J | W | z |
| Actual | a | 96% | 4% | 0% | 0% | 0% |
|  | b | 22% | 77% | 0% | 0% | 1% |
|  | j | 0% | 12% | 70% | 18% | 0% |
|  | w | 0% | 6% | 0% | 94% | 0% |
|  | z | 0% | 0% | 0% | 14% | 86% |

A set of 5 letters is chosen to test the performance of the DWT algorithm for the similar letters. The letters chosen are a, d, g, q and u. Again the tests were run 100 times for each letter and the average of the results are presented in Table IV.3.

Table IV.3: Detection accuracy of DTW algorithm for 5 similar letters. The size of the letters is 12 inches by 12 inches. The test is done 100 times for each letter.

|  |  | Predicted | | | | |
|---|---|---|---|---|---|---|
|  |  | a | d | g | q | u |
| Actual | a | 54% | 0% | 19% | 6% | 11% |
|  | d | 0% | 100% | 0% | 0% | 0% |
|  | g | 0% | 17% | 18% | 65% | 0% |
|  | q | 3% | 0% | 16% | 81% | 0% |
|  | u | 0% | 0% | 0% | 0% | 100% |

As expect, differentiating similar letter is more challenging than differentiating non-similar letters.

Table IV.4 provides the results for the entire English alphabet. The tests are run 20 times for each letter. The average accuracy is 71%. With the help of a spellchecker application, this can be dramatically improved.

## V. CONCLUSION

We presented a new approach to mobile text entry for wrist worn wearable systems, such as the popular smart watch. Motion sensors on the devices are used to extrapolate air-written letters. The data filtering and classification approaches are considered and evaluated on the AirDraw system prototype. An average 71% accuracy rate is found classifying all the letters of the English alphabet, with half the letters achieving at least 80% accuracy. In conclusion, air writing with a smart watch is shown to be a feasible and promising mobile user interface option for wrist worn wearable systems.

Table V.7: Accuracy percentage of DTW algorithm for the detection of lowercase English alphabet. The test is repeated 20 times for each letter. The size of the letters is 12 inches by 12 inches.

| | | Predicted | | | | | | | | | | | | | | | | | | | | | | | | | |
|---|---|---|---|---|---|---|---|---|---|---|---|---|---|---|---|---|---|---|---|---|---|---|---|---|---|---|---|
| | | a | b | c | d | e | f | g | h | i | j | k | l | m | n | o | p | q | r | s | t | u | v | w | x | y | z |
| Actual | a | 90 | | | | | | | | | | | | | | | | | | 10 | | | | | | | |
| | b | | 80 | | | | | 20 | | | | | | | | | 10 | | | | | | | | | | |
| | c | | | 60 | | | | | | | | 5 | | | 35 | | | | | | | | | | | | |
| | d | | | | 60 | | | 15 | 5 | | 5 | | | | | | | | | 5 | | | | | | 10 | |
| | e | | | | | 45 | 10 | | | | | | | | | | | | | 15 | | | | | 15 | | 15 |
| | f | | | | | | 85 | | | | | 15 | | | | | | | | | | | | | | | |
| | g | | | | | | 20 | 80 | | | | | | | | | | | | | | | | | | | |
| | h | | | | | | | | 80 | | | | | | 5 | | | | | | 10 | 15 | | | | | |
| | i | | | | | | | | | 95 | | | | | | | | | | | | 5 | | | | | |
| | j | | | | | | | | | 50 | 45 | | | | | | | | | | 5 | | | | | | |
| | k | | | | | | | 5 | 35 | | | 60 | | | | | | | | | | | | | | | |
| | l | | | | | | | | | | | | 85 | | | | | | | | | 15 | | | | | |
| | m | | | | | | | | | | | | | 85 | | | | | | 15 | | | | | | | |
| | n | | | | | | | | | | | | | | 60 | | | | | 10 | 30 | | | | | | |
| | o | | | | | | | | | 5 | 5 | | | | 80 | 5 | | | 5 | | | | | | | | |
| | p | | 5 | | | | | | | | | | | | | 5 | 85 | | | 5 | | | | | | | |
| | q | 15 | | 5 | | | 35 | | | | | | | | | | | 45 | | | | | | | | | |
| | r | | 5 | | | | | | | | | | | | | | | | 80 | 10 | | 5 | | | | | |
| | s | | 5 | 25 | | | 5 | 5 | | | | 5 | | | | | | | | 55 | | | | | | | |
| | t | | | | | | | 10 | 5 | | | | | | | | | | 5 | 80 | | | | | | | |
| | u | | | | 15 | 5 | | | | | | | 35 | | | | | | | 45 | | | | | | | |
| | v | | | | | | | 20 | | | | | | | | | | 35 | | | 55 | | | | | | |
| | w | | | | | | | | | | | | | | | | | | | | | | 100 | | | | |
| | x | | | | | | | | | | | | | | | 40 | | | | | | | 60 | | | | |
| | y | | | | 15 | | 10 | 5 | | | | | 5 | | | | | | | 5 | | | 10 | 50 | | | |
| | z | | | | | | | | 15 | | | | | | | | | | | | | | | | | 85 | |